\documentclass[lettersize,journal]{IEEEtran}
\usepackage{amsmath,amsfonts}
\usepackage{algorithmic}
\usepackage{algorithm}
\usepackage{array}
\usepackage[caption=false,font=normalsize,labelfont=sf,textfont=sf]{subfig}
\usepackage{textcomp}
\usepackage{stfloats}
\usepackage{url}
\usepackage{verbatim}
\usepackage{graphicx}
\usepackage{gensymb}
\usepackage{siunitx}
\usepackage{amssymb}
\usepackage{mathrsfs}
\usepackage{esvect}
\usepackage{bbding}
\usepackage{soulutf8}
\soulregister{\cite}7 
\soulregister{\ref}7 
\soulregister{\eqref}7
\usepackage{color}
\usepackage{multirow}
\usepackage{booktabs} 
\usepackage{diagbox} 
\usepackage{pdfpages}
\usepackage[numbers,sort&compress]{natbib}
\usepackage[colorlinks,linkcolor=blue,anchorcolor=blue,citecolor=blue]{hyperref}
\usepackage{xcolor}
\usepackage{tcolorbox}
\usepackage{tikz}
\usetikzlibrary{shapes.geometric}

\newcommand{\mytikzstar}[1][]{\begin{tikzpicture}[baseline={([yshift=-2pt]current bounding box.center)}]
  \def\radius{0.17cm}
  \draw[line width=0.1pt]
    (90:\radius) -- (234:\radius) -- (18:\radius) -- (162:\radius) -- (306:\radius) -- cycle;
\end{tikzpicture}
}

\newcommand\highlight[1]{\hl{#1}}

\newcommand\highlightReference[1]{%
  \expandafter\newcommand\csname highlightReference-#1\endcsname{}%
}
\let\oldbibitem\bibitem
\def\bibitem#1 #2\par{%
  \expandafter\ifx\csname highlightReference-#1\endcsname\relax
    \oldbibitem{#1}#2\par
  \else
    \oldbibitem{#1}\highlight{#2}\par
  \fi
}
\usepackage{lineno} 
\usepackage[colorlinks,linkcolor=blue,anchorcolor=blue,citecolor=blue]{hyperref}
\begin{document}

\bstctlcite{IEEEexample:BSTcontrol}

\title{\LARGE \bf 
Design, Dynamic Modeling and Control of a 2-DOF \\ Robotic Wrist Actuated by Twisted and Coiled Actuators}

\author{Yunsong Zhang, Xinyu Zhou, and Feitian Zhang\textsc{*}
\thanks{Yunsong Zhang and Feitian Zhang are with the College of Advanced Manufacturing and Robotics, and the State Key Laboratory of Turbulence and Complex Systems, Peking University, Beijing, 100871, China (emails: {zhangyunsong@stu.pku.edu.cn} and {feitian@pku.edu.cn}).}
\thanks{Xinyu Zhou is with the Department of Electrical and Computer Engineering, Michigan State University, East Lansing, MI 48824 USA (email: {zhouxi63@msu.edu}).}
\thanks{* Send all correspondences to F. Zhang.}}

\markboth{}%
{Shell \MakeLowercase{\textit{et al.}}: A Sample Article Using IEEEtran.cls for IEEE Journals}


\maketitle
\begin{abstract}
Artificial muscle-driven modular soft robots exhibit significant potential for executing complex tasks. However, their broader applicability remains constrained by the lack of dynamic model-based control strategies tailored for multi-degree-of-freedom (DOF) configurations. This paper presents a novel design of a 2-DOF robotic wrist, envisioned as a fundamental building block for such advanced robotic systems. The wrist module is actuated by twisted and coiled actuators (TCAs) and utilizes a compact 3RRRR parallel mechanism to achieve a lightweight structure with enhanced motion capability. A comprehensive Lagrangian dynamic model is developed to capture the module's complex nonlinear behavior. Leveraging this model, a nonlinear model predictive controller (NMPC) is designed to ensure accurate trajectory tracking. A physical prototype of the robotic wrist is fabricated, and extensive experiments are performed to validate its motion performance and the fidelity of the proposed dynamic model. Subsequently, comparative evaluations between the NMPC and a conventional PID controller are conducted under various operating conditions. Experimental results demonstrate the effectiveness and robustness of the dynamic model-based control approach in managing the motion of TCA-driven robotic wrists. Finally, to illustrate its practical utility and integrability, the wrist module is incorporated into a multi-segment soft robotic arm, where it successfully executes a trajectory tracking task.

\end{abstract}

\begin{IEEEkeywords}
Robotic wrist, modular soft robots, dynamic modeling, soft actuator applications, twisted and coiled actuators.
\end{IEEEkeywords}

\section{Introduction}

\IEEEPARstart{T}{he} wrist serves as a key joint for facilitating the dexterous and agile movement in the human hand system \cite{bajaj2019state}. Inspired by this biological mechanism, robotic wrists have been extensively studied and successfully deployed in industrial manipulators and humanoid robots for manipulation and grasping tasks. Traditional robotic wrists are typically actuated by multiple motors, each controlling a single degree-of-freedom (DOF) rotational motion, forming serial wrist configurations \cite{bajaj2019state}. However, these conventional designs are often characterized by substantial mass and inertia, primarily due to the embedded mechanical components---such as DC motors and gearboxes--- within the joint structure. This limitation becomes particularly pronounced in emerging paradigms such as multi-segment robotics, where multiple joints are serially connected to form articulated manipulators \cite{bai2025kinematic, Bian2025accurate, liu2023inchworm, cheng2019design}. In such systems, the cumulative mass and inertia introduce by conventional modules can quickly become a bottleneck, severely impairing the system's dynamic responsiveness and energy efficiency. Consequently, the development of lightweight, compact, and high-performance joint modules is not simply advantageous---it is imperative for the continued advancement of modular and soft robotic platforms.

To address these limitations, increasing attention has been directed toward integrating artificial muscles into robotic systems \cite{liu2023inchworm, cheng2019design}. These actuators offer high energy density, low weight, and compact form, rendering them particularly suitable for advanced robotic applications. Their integration often results in substantial weight reduction while enabling smooth, biomimetic motion---characteristics especially desirable for tasks involving human-robot interaction or delicate manipulation. Among the most commonly employed artificial muscles are shape memory alloys (SMAs) \cite{salerno2016novel,hyeon2023lightweight,copaci2020sma} and dielectric elastomer actuators (DEAs) \cite{li2018bioinspired,xing2020super}. Several multi-DOF robots have been developed based on these technologies.. For instance, Salerno \textit{et al.} \cite{salerno2016novel} proposed an origami-based gripper actuated by SMAs, while Hyeon \textit{et al.}\cite{hyeon2023lightweight} designed a prosthetic wrist  leveraging SMA actuation. In parallel, Li \textit{et al.} \cite{li2018bioinspired} utilized DEAs to develop robotic eyeballs, and Xing \textit{et al.} demonstrated a lightweight, soft manipulator actuated by DEAs \cite{xing2020super}. Despite their advantages, each type of artificial muscle exhibits inherent trade-offs. SMAs can generate relatively large forces, but are limited by low actuation frequencies and pronounced nonlinearities \cite{seok2012meshworm,lin2011goqbot}. DEAs, on the other hand, exhibit rapid and precise activation, but typically provide low force outputs and demand high actuation voltages, often exceeding 10 kV \cite{acome2018hydraulically,jung2007artificial}.

The twisted-and-coiled actuator (TCA) provides a promising alternative for robotic wrist design. TCAs, characterized by their helical structures formed from twisted nylon threads or spandex fibers, produce axial contraction or elongation in response to thermal excitation, governed by their structural parameters. TCAs exhibit several appealing attributes, including large contraction ratios, high force output, substantial load capacity, impressive energy density, and easy of control under low-voltage operation \cite{haines2014artificial,zhang2019robotic, zhou2022force, horton2019consistent}. These features position TCAs as strong candidates for  compact actuator integration, supporting the development of highly dense and responsive mechatronic systems.

To enable precise actuation of TCAs and expand their applicability in robotic systems, researchers have conducted extensive studies to characterize their properties and dynamic behaviors \cite{masuya2017nonlinear, bao2023fast, yip2017control, karami2020modeling, zhang2020modeling}. A key challenge lies in the intrinsic nonlinearity of TCAs, which emerges from the coupled thermal and mechanical dynamics. The thermal response is predominantly governed by Joule heating, while natural convection serves as the primary mode of heat dissipation. Additional complexities arise from  radiative losses and heat damping effects \cite{masuya2017nonlinear, bao2023fast}. On the mechanical side, TCA behavior is also nonlinear. While force output is often approximated via a linear model incorporating stiffness, damping, and temperature effects \cite{yip2017control}, more comprehensive models capture temperature-induced variations in thermal expansion and elastic modulus \cite{karami2020modeling} as well as hysteresis phenomena \cite{zhang2020modeling}. Despite these nonlinearities, linear dynamic models \cite{yip2017control} remain prevalent due to their tractability, effectiveness in closed-loop control, and ease of parameter identification, often achieving comparable performance in motion control tasks \cite{wang2021lightweight, hu2024axial}.

Researchers have utilized TCAs to develop multi-DOF robotic systems \cite{sun2023development, yang2020compact, yang2024variable, wu2018biorobotic, wu2018novel}. Examples include the minimally invasive surgical robot by Sun \textit{et al.} \cite{sun2023development}, the variable-stiffness soft robotic manipulator by Yang \textit{et al.} \cite{yang2024variable}, and the modular artificial musculoskeletal system proposed by Wu \textit{et al.} \cite{wu2018biorobotic}. However, a common limitation in these pioneering efforts is the reliance on simplified control strategies, typically kinematics-based PID controllers, which overlook the complex and coupled dynamics inherent to TCA systems  \cite{yang2024variable, sun2023development, yang2020compact}. While such approaches suffice for proof-of-concept demonstrations, they inherently limits the system's precision, responsiveness, and adaptability, especially in scenarios involving multi-module integration, where dynamic coupling and error propagation become increasingly significant \cite{liu2023inchworm, cheng2019design}. The lack of accurate dynamic models and corresponding model-based control strategies remains a fundamental bottleneck to achieving the full performance potential of TCA-driven robotic platforms.

To address the aforementioned challenges, this paper presents a novel 2-DOF parallel robotic wrist actuated by TCAs within a compact parallel mechanism. A Lagrangian dynamics model is rigorously established and subsequently validated through open-loop control experiments. Leveraging this model, a nonlinear model predictive controller (NMPC) is designed and evaluated through extensive trajectory tracking experiments, benchmarked against a baseline proportional-integral-derivative (PID) controller. To demonstrate the modularity and scalability of the proposed wrist, we further construct a multi-joint robotic system by serially connecting multiple wrist units, highlighting a viable pathway toward complex, high-DOF soft robotic architectures.

The contributions of this paper are threefold. First, building upon prior studies in parallel mechanisms \cite{kim2018quaternion,kim2015design}, we propose a novel design methodology for a TCA-actuated robotic wrist. The design enables precise and responsive motion within a compact mechatronic envelope. Second, a comprehensive Lagrangian dynamics model is derived and validated through extensive experimental evaluation. Based on the dynamic model, an NMPC is designed demonstrating superior trajectory tracking performance compared to conventional PID control, thus validating the benefits of model-based strategies in TCA systems.  Finally, we demonstrate the modularity and integration potential of the wrist by constructing a Multi-Segment Soft Robot Arm (MSRA) composed of three serially connected wrist modules. The MSRA successfully performs complex spatial trajectories, confirming the effectiveness of the wrist not only as a standalone actuator but also as a scalable building block for advanced modular robotic platforms.

\section{Design of Robotic Wrist}
\begin{figure}[h]
        \centering
        \includegraphics[width=0.5\textwidth]{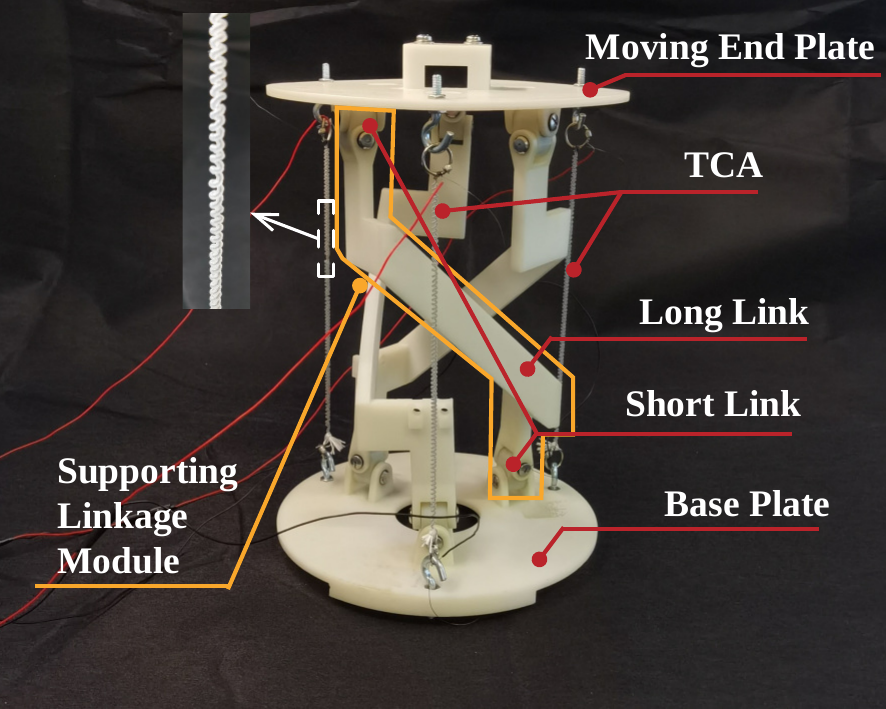}
        \caption{Illustration of the robotic wrist design and associated components. The wrist design consists of a base plate, a moving end plate, and three supporting linkage modules, each composed of a long link, two short links, and four joints. All three TCAs are securely fastened to the perimeters of the base and end plates.}
        \label{prototype}
\end{figure}

The utilization of parallel mechanisms in robotic wrist design has gained increasing attention in recent years \cite{kim2018quaternion,pang2022design,kuang2023haptic,wang2023modular}. Illustrated in Fig.~\ref{prototype}, the parallel wrist design features a compact structure engineered to avoid interference among supporting linkages while offering a broad range of motion without singularities. Tendons and cables have been successfully applied to actuate robotic wrists with parallel mechanisms \textcolor{black}{\cite{kim2018quaternion,kim2015design}}. Given their string-like nature, artificial muscles like TCAs are expected to achieve similar success. The wrist design consists of a base plate, a moving end plate, and three supporting linkage modules adopting an RRRR configuration \cite{merlet2006parallel}, each composed of a long link, two short links, and four joints, connecting the two plates. All three TCAs are securely fastened to the perimeters of the base and end plates, equally spaced with 120$^\circ$ separation between each pair of neighboring actuators. \par

\begin{figure}[h]
        \centering
        \includegraphics[width=0.5\textwidth]{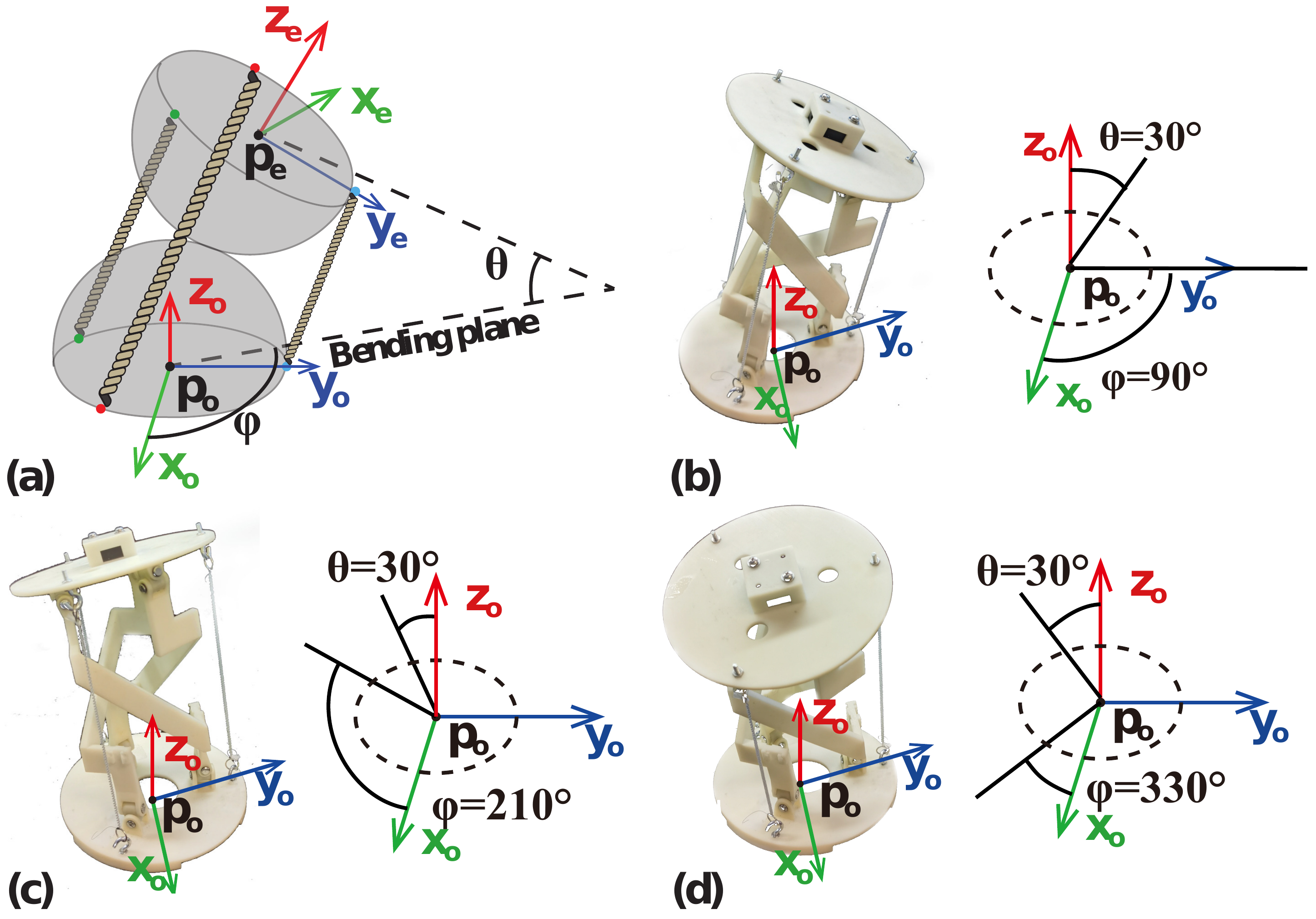}
        \caption{Illustration of the approximated motion model of the robotic wrist with its bending motion states. (a) The motion of the robotic wrist is modeled as the rolling motion of an end hemisphere relative to a stationary base hemisphere; (b), (c) and (d) the bending and swinging motions of the TCA-driven robotic wrist at bending directions $\varphi = 90^\circ, 210^\circ, 330^\circ$ with bending angle $\theta = 30^\circ$.}
        \label{Fig4}
\end{figure}

The motion of the wrist, specifically the rotational and translational movements of the end plate, is approximately modeled as the rolling motion of an end hemisphere relative to a stationary base hemisphere \cite{kim2018quaternion} as illustrated in Fig.~\ref{Fig4}. We define the base reference frame $\sum_o$, affixed to the base plate, denoted as $p_o-x_oy_oz_o$, and the end reference frame $\sum_e$, affixed to the end plate, denoted as $p_e-x_ey_ez_e$. A bending plane is defined, incorporating the $z$-axis $z_o$ of the base frame and the centerline $\vv{p_op_e}$, which connects the geometric centers of the base and end plates. The bending direction $\varphi$ and the bending angle $\theta$ form the complete set of motion states or the pose of the end plate, with $\varphi$ defined as the angle between the $x_o$ axis and the bending plane, and $\theta$ defined as the angle between $z_o$ and $z_e$ within the bending plane. By controlling the temperature of the TCAs, the robotic wrist is driven to specific positions, as illustrated by the bending and swinging motion postures in Fig.~\ref{Fig4}(b), (c), and (d).

\section{Dynamic Model}
\subsection{Review of TCA Model}
The TCA model adopted in this paper consists of two interconnected sub-models, including the thermoelectric model and the dynamics model \cite{yip2017control}. Specifically, the thermoelectric model of the TCA describes the physical process as a heat transfer phenomenon featuring a heat source within the actuator, approximated as a first-order model \cite{yip2017control}, i.e.,
\begin{equation}
        \begin{aligned}
                C_{\rm th} \frac{\mathrm{d}T}{\mathrm{d}t} = P(t)-\lambda (T(t)-T_{\rm amb}) 
                \label{thermodynamic}
        \end{aligned}        
\end{equation}
where $C_{\rm th}$ denotes the thermal mass of the actuator ($\rm Ws$/$^{\circ}$C), $P(t)$ represents the thermal power applied to the actuator at time $t$, $T(t)$ denotes the temperature of the actuator at time $t$, 
$T_{\rm amb}$ represents the ambient temperature, and $\lambda$ denotes the absolute thermal conductivity of the actuator (W/$^{\circ}$C).\par
The dynamics model of the TCA is equivalent to a spring-damper system supplemented by a heat power source, i.e.,
\begin{equation}
        \begin{aligned}
                F=k(L-L_0)+b\dot{L}+c(T-T_{\rm amb})
                \label{TCA_F}
        \end{aligned}
\end{equation}
Here, $F$ represents the tensile force generated by the TCA, while $b$, $k$ and $c$ denote the damping, spring stiffness, and temperature coefficients of the actuator, respectively. 
$L$ and $L_0$ denote the current length and the original length of the TCA, respectively.

\subsection{Kinematics Modeling of Robotic Wrist}
\subsubsection{Inverse Kinematics Modeling}
The inverse kinematics model determines the lengths of the TCAs $L_i$ given the robotic wrist's pose, characterized by the bending direction $\varphi$ and the bending angle $\theta$.
As illustrated in Fig.~\ref{Fig4}, the pose of the end frame $\sum_e$ relative to the base frame $\sum_o$ is calculated through a sequence of five elementary rotations and translations, including 
$\left. 1\right)$ rotating about the $z_o$ axis by $\varphi$, 
$\left. 2\right)$ rotating about the intermediate $y$-axis by $\frac{\theta}{2} $,
$\left. 3\right)$ moving forward along the positive direction of the $z$-axis by $h$, 
$\left. 4\right)$ rotating about the $y$-axis by $\frac{\theta}{2}$, and  
$\left. 5\right)$ rotating about the $z$-axis by $\varphi$. 
\par
The coordinate transformation matrix from the base frame to the end frame is then given by
\begin{align}
        &^oT_e = R_z(\varphi) R_y(\frac{\theta}{2})   T_z(h)   R_y(\frac{\theta}{2})  R_z(-\varphi) \nonumber \\
        &=\small\left[
                \begin{matrix}
                        1-2 C^2 \varphi S^2 \frac{\theta}{2} & - S 2\varphi S^2 \frac{\theta}{2} & C \varphi S \theta & h C \varphi S \frac{\theta}{2}\\
                        -S 2\varphi S^2 \frac{\theta}{2} & 1 - 2 S^2 \varphi S^2 \frac{\theta}{2} & S \theta S \varphi & h S \varphi S \frac{\theta}{2} \\
                        -C \varphi S \theta & -S \varphi S \theta & C \theta & h C \frac{\theta}{2}\\
                        0 & 0 & 0 &1
               \end{matrix}\right]
            \normalsize
\end{align} 
Here, $S(\cdot)$ and $C(\cdot)$ denote the sine and cosine functions, respectively, while $h$ denotes the distance between $p_o$ and $p_e$, or equivalently, the diameter of the rolling hemisphere. $R_z$, $R_y$, and $T_z$ represent the corresponding homogeneous transformations for rotations about the $z$ and $y$ axes and translation along the $z$ axis, respectively.\par 
The lengths of the TCAs are calculated as
\begin{align}
        \left\lvert L_1 \right\rvert &=\left\lVert ^oT_e P^1_{e}-P^1_{o} \right\rVert = h-2 r S (\varphi) S (\frac{\theta}{2}) \label{4} \\
        \left\lvert L_2 \right\rvert &=\left\lVert ^oT_e P^2_{e}-P^2_{o} \right\rVert = h-2 r S (\varphi-\frac{2\pi}{3}) S (\frac{\theta}{2}) \label{5} \\ 
        \left\lvert L_3 \right\rvert &=\left\lVert ^oT_e P^3_{e}-P^3_{o} \right\rVert = h-2 r S (\varphi+\frac{2\pi}{3}) S (\frac{\theta}{2}) \label{6}
\end{align}
where $P^i_{e}$ and $P^i_{o}$ represent the coordinates of the connection points of the $i$-th TCA with the end plate and the base plate, respectively, expressed in the reference frames $\sum_e$ and $\sum_o$.
$r$ denotes the radius of the base plate. \par
A constraint relationship governs the lengths of the three TCAs, i.e., 
\begin{align}
        \left\lvert L_1 \right\rvert+ \left\lvert L_2 \right\rvert+\left\lvert L_3 \right\rvert=3h
\label{L-relationship}
\end{align}

\subsubsection{Forward Kinematics Modeling}
The forward kinematics model determines the pose of the wrist, i.e., the bending direction $\varphi$ and the bending angle $\theta$, given the lengths of the TCAs $L_i$ where $i\in\{1,2,3\}$.  \par
Under the constraint on TCA lengths (Eq.~(\ref{L-relationship})), the bending angle $\theta$ and bending direction $\varphi$ are determined by solving Eqs.~(\ref{4})--(\ref{6}), resulting in
\begin{align}
        \theta &=2\arcsin(\frac{\sqrt{-L_1^2-L_2^2-L_3^3+L_1L_2+L_1L_3+L_2L_3}}{3r}) \\
        \varphi &= \arcsin(\frac{-2L_1+L_2+L_3}{2 \sqrt{-L_1^2-L_2^2-L_3^2+L_1L_2+L_1L_3+L_2L_3}})
\end{align}
\par
\begin{figure*}[htbp]
        \begin{small}

        \begin{align}
                &^oR_e =T_y(\delta_1,0,r,0) T_x(\delta_2+\delta_{0},0,0,l_1)  T_x(\delta_2-\delta_{0},0,0,l_2) T_y(\delta_1,0,r,l_1) \nonumber \\
                &=\left[ \begin{matrix}                     1-2S^2(\delta_1)C^2(\delta_2) & S(\delta_1)S(2\delta_2)& S(2\delta_1)C^2(\delta_2) & S(\delta_1)(l_1 + l_1C(2\delta_2) + 2rS(2\delta_2) + l_2C(\delta_2 + \delta_{0}))\\                S(\delta_1)S(2\delta_2) & C(2\delta_2) &   -C(\delta_1)S(2\delta_2) &r + 2rC(2\delta_2) - l_1S(2\delta_2) - l_2S(\delta_2 + \delta_{0})\\                 -S(2\delta_1)C^2(\delta_2) & C(\delta_1)S(2\delta_2) &  2S^2(\delta_1)C^2(\delta_2) + C(2\delta_2) & C(\delta_1)(l_1 + l_1C(2\delta_2) + 2rS(2\delta_2) + l_2C(\delta_2 + \delta_{0}))\\                0&0&0&1                 \end{matrix}                \right]
                \label{forward-trans}
        \end{align}
        \noindent\rule{\textwidth}{0.4pt}

        \end{small}
        \end{figure*}

Furthermore, we investigate the relationship between the joint angles and the pose of the wrist, thereby facilitating subsequent analysis of wrist dynamics.

\begin{figure}[htbp]
        \centering
        \includegraphics[width=0.5\textwidth]{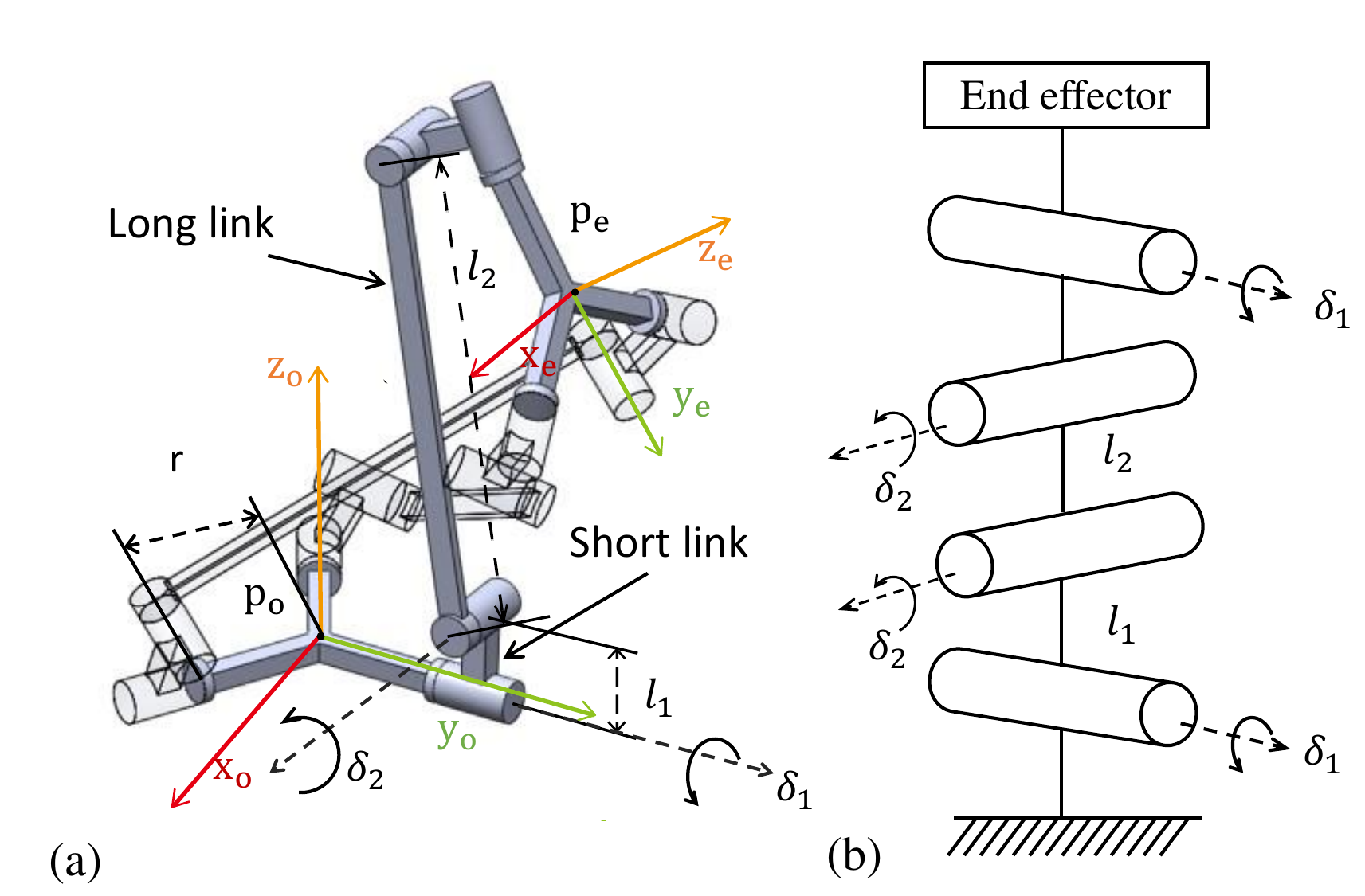}
        \caption{The linkage model of the robotic wrist driven by TCAs.  (a) The CAD model of the linkage configuration that consists of two short links, one long link, and four joints, and (b) the effective linkage model as an open chain mechanism with four rotating joints. }
        \label{Fig5}
\end{figure}

Figure~\ref{Fig5} illustrates the equivalent linkage model of the wrist, incorporating the constraint on joint angles during motion. Define $l_1$ as the length of the short link, $l_2$ as the length of the long link, and $\delta_{0}$ as the default angle between the long and short links when all the TCA actuators are at rest and the end plate is horizontal. The forward rotation translation matrix is obtained by Eq.~(\ref{forward-trans}) where $\delta_1$ and $\delta_2$ denote the joint angle between the short link and the connecting base/end plate, and the joint angle between the short and long links, respectively.

The geometric relationship between the length of the long link $l_2$ and the radius of the base plate $r$ reads
\begin{align}
        l_2\sin(\delta_{0})=2r
\end{align}

\par The joint angles and the wrist' pose follow geometric constraints, i.e.,
\begin{align}
        \delta_{1} &= \arctan (-\cos \varphi  \tan \frac{\theta}{2} )\\
        \delta_{2} &= \arctan ( \tan \varphi  \sin \theta) 
\end{align}

\par

The three supporting linkages in the robotic wrist exhibit rotational symmetry, with a rotational angle of $120^\circ$, and the center of rotation is located at the center $p_o$ of the base plate. Consequently, we express the joint angles of other two linkages as 
\begin{align}
        \delta_{2i+1} &= \arctan (-\cos (\varphi+\frac{2i\pi}{3})  \tan \frac{\theta}{2} )\\
        \delta_{2i+2} &= \arctan ( \tan (\varphi+\frac{2i\pi}{3})  \sin \theta),   \quad  (i=1,2)
\end{align}
where $\delta_{2i+1}$ represents the joint angle between the short link and the connecting base/end plate, and $\delta_{2i+2}$ represents the corresponding joint angle between the short and long links.

\subsection{Dynamic Modeling of Robotic Wrist}
Given the complexity of the internal forces and the presence of multiple linkages within the parallel robotic wrist, the Lagrangian modeling method offers a more effective approach compared to the Newton-Euler method. Considering the pose of the robotic wrist is described by the bending angle $\theta$ and the bending direction $\varphi$, 
we set the generalized coordinates as $q=[\theta ,\varphi]$ and $\dot{q}=[\dot{\theta},\dot{\varphi}]$. 
The kinetic energy of the robotic wrist is expressed as
\begin{align}
        T(q,\dot{q},t)=T_{0}(q,\dot{q},t)+\sum_{i=1}^{3}T_{i}(q,\dot{q},t)
\end{align}
Here, $T$ represents the total kinetic energy of the wrist in motion, 
$T_{0}$ denotes the kinetic energy of the end plate,
and $T_{i}$ denotes the kinetic energy of the $i$-th supporting linkage. 
Specifically, $T_{0}$ is calculated as
\begin{align}
      &T_{0}=\frac{1}{8} M \dot{q} 
        \small\left[
        \begin{matrix}
             h^2+r^2 & 0\\
             0 & r^2(1+\cos^2{\theta})+4h^2 \sin^2(\frac{\theta}{2})
        \end{matrix}
       \right]\small
       \dot{q}^T
\end{align}
where $M$ represents the mass of the end plate, $h$ denotes the distance between $p_o$ and $p_e$. 
$T_i$ is calculated as
\begin{align}
        &T_{i}(q,\dot{q},t)=\frac{1}{8} m l_2^2 \omega_i^T \left[
                \begin{matrix}
                        0 & 0 & 0\\
                        0 & 1 & 0 \\
                        0 & 0 & 1
                \end{matrix}
        \right] \omega_i 
        +\small\frac{1}{2}\omega_i^T
            \mathbf{I}_m 
            \omega_i
\end{align}
where $m$ and $\mathbf{I}_m$ represent the mass and inertia of each supporting linkage, respectively, while $\omega_i$ denotes the angular velocity of the $i$-th linkage with respect to the base frame $\sum_o$, given by
\begin{align}
        \omega_i&=\left[
                \begin{matrix}
                \dot \delta_{2i-1} S(\delta_{0}-\delta_{2i}) &- \dot \delta_{2i-1} C(\delta_{0}-\delta_{2i}) &\dot \delta_{2i}
                \end{matrix}
            \right]^T
\end{align}

The total potential energy is expressed as
\begin{align}
        V(q,t)=M g h \cos \frac{\theta}{2}+ \sum_{i=1}^3(\frac{l_2}{2} m g \cos(\delta_{0}-\delta_{2i}))
\end{align}

The Lagrangian of the robotic wrist system $\mathcal{L}(q,\dot{q},t)$ is then given by
\begin{align}
        \mathcal{L}(q,\dot{q},t)=T(q,\dot{q},t)-V(q,t)
\end{align}
\par Finally, the Lagrangian dynamics model is derived in the generalized coordinate system $q=[\theta ,\varphi]$, i.e.,
\begin{align}
        \frac{d }{dt}(\frac{\partial \mathcal{L}}{\partial \dot q })-\frac{\partial \mathcal{L}}{\partial q}
        &=\left[
                \begin{matrix}
                        Q_{\theta}\\Q_{\varphi} 
                \end{matrix}
        \right]
        \label{dynamic_model}
\end{align}
where $Q_{\theta}$ and $Q_{\varphi}$ denote the generalized forces acting on the robot associated with $\theta$ and $\varphi$, respectively, collectively forming the generalized force vector $Q=[Q_{\theta},\,Q_{\varphi}]^T$.

We derive the dynamic equation of the robotic wrist, i.e.,
\begin{align}
           M(q) \ddot{q} + V(q, \dot{q}) + D(q) \dot{q} + G(q) = Q(q, \dot{q}, T)
                \label{dynamic_eq}
          \end{align}
where $M(q) \in \mathbb{R}^{2 \times 2}$, $V(q, \dot{q}) \in \mathbb{R}^{2}$, $D(q) \in \mathbb{R}^{2 \times 2}$, $ G(q) \in$ $\mathbb{R}^{2}$, and $Q(q, \dot{q}, T) \in \mathbb{R}^{2} $ are the inertia matrix, the centrifugal and Coriolis term, the damping term, the gravitational term, and the generalized force, respectively.

Assuming that the frictional torque at the bearing is negligible during the rotation process, the $j$-th component of the generalized force vector $Q_j$ is calculated as
\begin{align}
        Q_j= \sum_{i=1}^3 F_i  \frac{\partial L_i}{\partial q_j} 
\end{align}
where $F$ represents the active force matrix, calculated as a function of the TCA model.

\section{Control Design}
\subsection{Control-Oriented State-Space Robot Dynamics}

This section derives the robot's dynamic equations utilized in the NMPC design, incorporating Eqs.~$\eqref{thermodynamic}$, $\eqref{TCA_F}$, and $\eqref{dynamic_eq}$.

In this paper, the generalized coordinates of the robotic wrist, the corresponding coordinate velocities, and the temperature of the TCA are considered as system state variables, denoted as $x = [q, \dot{q}, T]^T \in \mathbb{R}^{7}$. The electrical power input at both ends of the TCA is defined as the system input $u  = [P_1, P_2, P_3] \in \mathbb{R}^{3}$ where $P_i=I_i^2R_i$, with $I_i$ and $R_i$ representing the current and the resistance of the $i$-th TCA. Power consumption occurs primarily within the TCAs with negligible losses attributed to the electrical wiring. Temperature of TCAs $\mathit{T}\in \mathbb{R}^3$ is not directly observed but calculated within the controller using the thermodynamic model Eq.~(\ref{thermodynamic}).

The dynamics of the robotic wrist for control design is then expressed in the state space, i.e.,
{\small
\begin{align}
    \dot{\mathit{x}}&=\left[
    \begin{matrix}
         \dot{q} \\ \ddot{q}\\ \dot{\mathit{T}}
    \end{matrix}
    \right]=\left[
    \begin{matrix}
        \dot{q} \\
         M(q)^{-1}(Q(q, \dot{q}, T),- V(q, \dot{q}) - D(q) \dot{q} - G(q))\\
        \dfrac{\mathit{u}-\lambda(\mathit{T}-\mathit{T_0})}{C_{th}}
    \end{matrix}
    \right]
\end{align}
}

\subsection{Nonlinear Model Predictive Control}
This section introduces a nonlinear model predictive controller (NMPC) designed for the robotic wrist. The controller aims to accurately track predefined trajectories within the motion space while constraining input to achieve low energy consumption, thereby maximizing the driving potential of the TCA. Since the optimization process of NMPC is generally executed in discrete time domains, we discretize the derived dynamics via the Newton-Euler method, i.e. \cite{hanover2021performance}:
\begin{align}
        &\mathit{x}_{k+1}=\mathit{f}(\mathit{x}_{k}, \mathit{u}_{k})
\end{align}

 NMPC comprehensively considers the system dynamics and constraints, leveraging this method within the prediction horizon to compute the optimal control action. Specifically, the NMPC for the robotic wrist is formulated as  an optimization problem, i.e.,
\begin{align}
        &\min_{\mathit{u}_k^*}  \sum_{k=i}^{i+\mathit{N}-1}
        \left(\Vert \mathit{q}_k-\mathit{q}_k^r \Vert _\mathit{Q}^2 + \Vert \mathit{u}_k-\mathit{u}_{k-1} \Vert_\mathit{R}^2\right)
        +\Vert \mathit{u}_{k} \Vert_\mathit{S}^2 \nonumber \\
        &\text{s.~t.}~~~ \mathit{x}_{k+1}=\mathit{f}(\mathit{x}_k, \mathit{u}_k) ,\,x_0=x_{\text{init}}\nonumber \\
        &~~~~~~~~ \mathit{u}_{\min} \leq \mathit{u}_k \leq \mathit{u}_{\max},\,k=i,\dots,i+\mathit{N}-1 \nonumber \\
        &~~~~~~~~ 0 \leq i \leq p-1
        \label{eq:cost}
\end{align}
where $\mathit{N} $ represents the control horizon length, $ \mathit{u}_k $ denotes the input at time instant $k$ where $k$ spans from time $ i $ to $ i+\mathit{N}-1 $. The parameter $ p $ signifies the prediction horizon length, $ \mathit{q}_k $ is the predicted generalized coordinates, and $ \mathit{q}_k^r $ represents the reference trajectory, evaluated for time instant $ k $. The matrices $ \mathit{Q} $, $ \mathit{R} $, and $ \mathit{S} $ are positive-definite weighting matrices for the state, power input variation, and power input magnitude, respectively. The NMPC problem is solved using the sequential quadratic programming (SQP) method, based on the direct multiple shooting approach \cite{johansen2011introduction}.
\begin{figure}[htbp]
    \centering
   \includegraphics[width=0.5\textwidth]{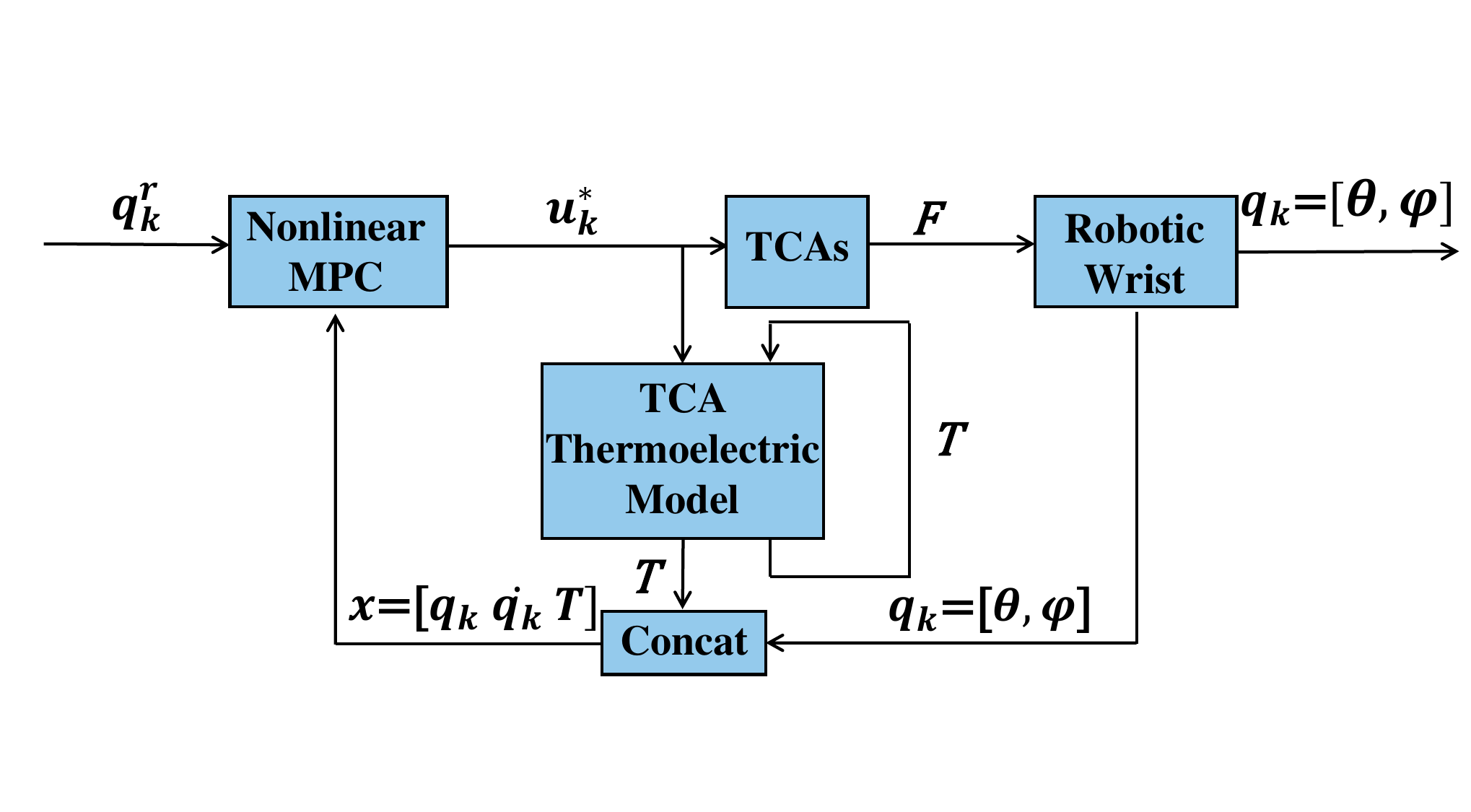}
    \caption{Block diagram of the NMPC for the robotic wrist.}
    \label{control_box}
\end{figure}

Figure~\ref{control_box} illustrates the control framework of the NMPC for the robotic wrist. The reference trajectory $q^r_k$ is input into the NMPC, which calculates the optimal power input $u^{\text{*}}_k$ for the TCA. The TCA, exerting an output force on the wrist, results in pose $q$, which is measured via an IMU. Alongside, the TCA temperature $T$, computed from the thermodynamic model, is fed back. This integrated feedback is used by the NMPC to optimize control actions continuously.

\section{Experiment}
\subsection{Experimental Setup}
\label{sec:exp_setup} 
\begin{figure}[h]
        \centering
        \includegraphics[width=0.5\textwidth]{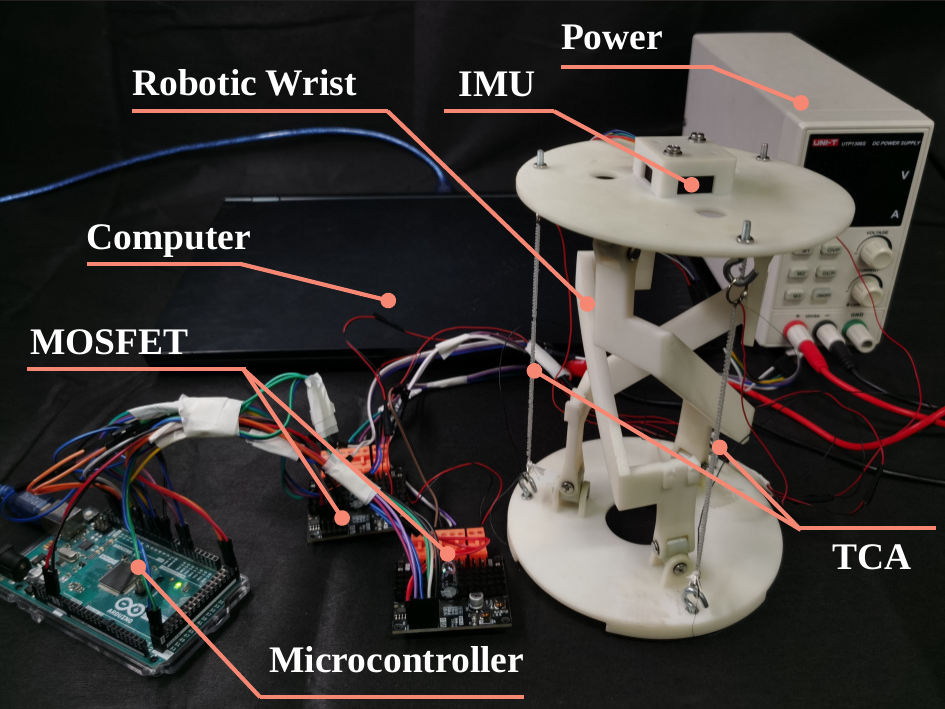}
        \caption{Experimental setup of the robotic wrist.}
        \label{hardware}
\end{figure}

The experimental setup of the robotic wrist comprises four main components including the computer, microcontroller, driver-sensor module, and TCAs, illustrated in Fig.~\ref{hardware}. The computer communicates with the microcontroller (Arduino Mega2560) via a serial port, leveraging the microcontroller to receive sensor output signals and control the TCAs. An inertial measurement unit (IMU) (N100MINI, Wheeltec) is employed to measure the pose of the end plate in real time. Once the microcontroller receives sensor measurements, it adjusts the output power of the MOSFETs, and controls TCAs to attain the desired temperature, thereby regulating the movement of the robotic wrist. The robotic wrist in this paper has a length of 170 mm and a maximum diameter of 90 mm. The parameters of the robotic wrist, TCAs, and NMPC utilized in this study are detailed in Table~\ref{para}.

\begin{table}[htbp] 
    \centering 
    \caption{Parameters of the robotic wrist and NMPC design.}        
    \label{para}
    \begin{tabular}{ccc} 
        \toprule 
        \textbf{Symbol} & \textbf{Description} & \textbf{Value} \\ 
        \midrule 
        $k$ & Spring constant of TCA & 238 \text{ N/m} \\ 
        \hline 
        $b$ & Damping coeff of TCA & 0.61 \text{ N·s/m} \\ 
        \hline 
        $c$ & Thermal constant of TCA & 23.09 \text{ mN/$^\circ$C} \\ 
        \hline 
        $R$ & Electrical resistance of TCA & 20 \text{$\Omega$} \\ 
        \hline 
        $C_{th}$ & Thermal mass of TCA & 0.8236 \text{ Ws/$^\circ$C} \\ 
        \hline 
        $\lambda$ & Thermal conductivity of TCA & 0.0235 \text{ W/$^\circ$C} \\ 
        \hline 
        $T_{amb}$ & Ambient temperature & 25 \text{ $^\circ$C} \\ 
        \hline 
        $L_0$ & Original length of TCA & 100 \text{ mm} \\ 
        \hline 
        $M$ & Mass of the end plate & 70 \text{ g} \\ 
        \hline 
        $h$ & Distance between base and end plate & 150 \text{ mm} \\ 
        \hline 
        $r$ & Radius of end plate & 50 \text{ mm} \\ 
        \hline 
        \multirow{2}{*}{$\mathbf{I}_m $} & \multirow{2}{*}{Inertia of supporting linkage} &  diag(82, 0.1, 82) \\ 
        & & \text{ kg*m$^2$} \\
        \hline
        $m$ & Mass of supporting linkage & 30 \text{ g} \\ 
        \hline 
        $l_2$ & Length of the long link & 180 \text{ mm} \\ 
        \hline 
        $g$ & Gravitational acceleration & 9.8 \text{ m/s$^2$} \\ 
        \hline 
        $Q$ & State weighting matrix & $\text{diag}(25, 25)$ \\ 
        \hline 
        $R$ & Input variation weighting  & $\text{diag}(2, 2, 2)$ \\ 
        \hline 
        $S$ & Input magnitude weighting  & $\text{diag}(0.25, 0.25, 0.25)$ \\ 
        \bottomrule 
    \end{tabular} 
\end{table} 

\subsection{Model Validation of Robotic Wrist Dynamics}

\begin{figure}[htbp]
        \centering
        \includegraphics[width=0.5\textwidth]{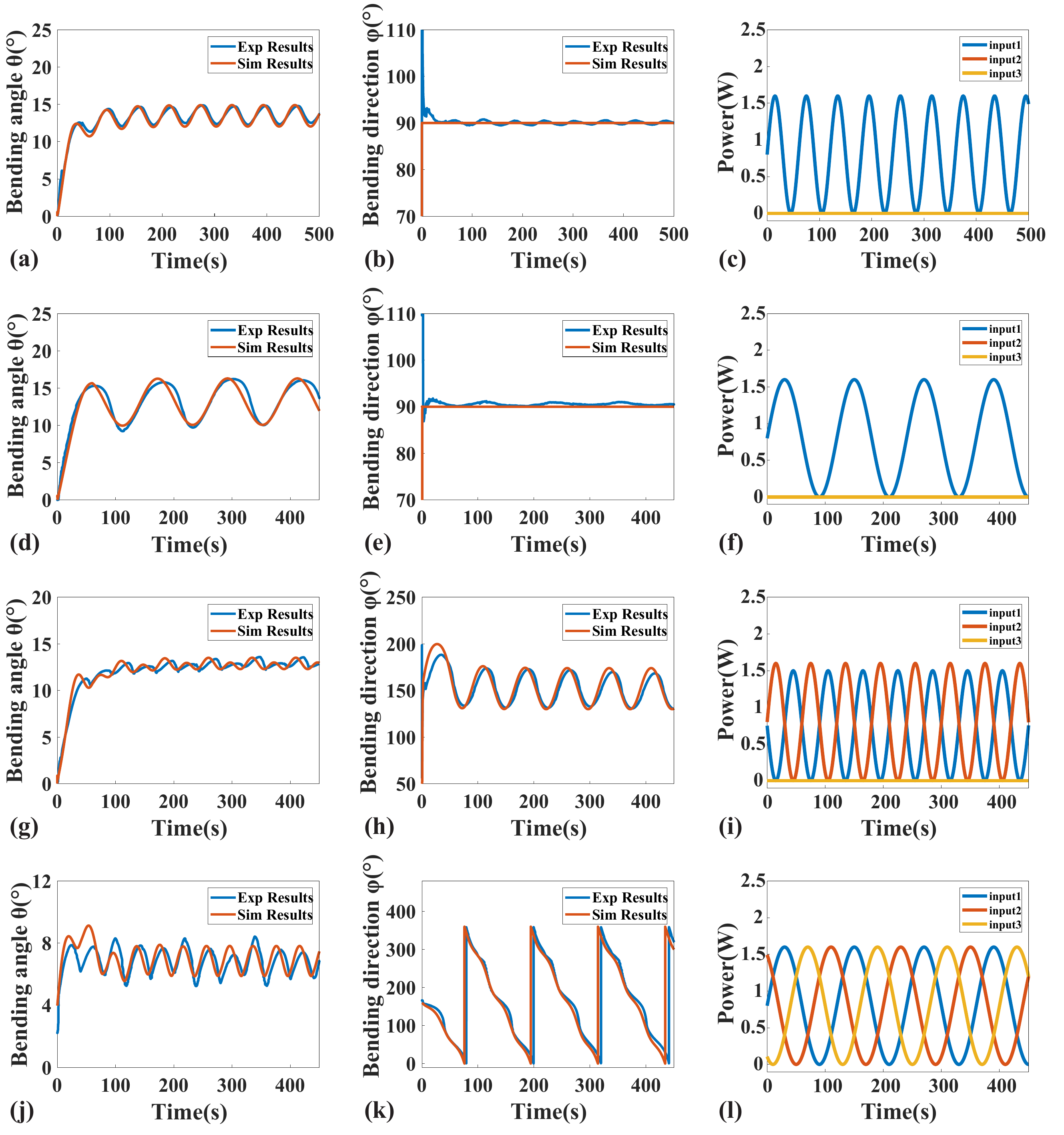}
        \caption{Model validation experimental results under sinusoidal control inputs. The first and second rows illustrate the comparison between experimental results and model predictions for the bending angle $\theta$ (a \& d) and bending direction $\varphi$ (b \& e) under a control input (c \& f) where only one single TCA is actuated, following a sinusoidal function with frequencies of 1/60\,Hz and 1/120\,Hz, respectively. The third and fourth rows show a similar comparison for the bending angle $\theta$ (g \& j) and bending direction $\varphi$ (h \& k) under a control input (i \& l) with two and three TCAs actuated, respectively.}
        \label{sinmodelval}
\end{figure}

We validated the derived dynamics model \textcolor{black}{Eq.~(\ref{dynamic_eq})} through  extensive open-loop experiments using a range of representative control inputs, and compared the experimental results against model predictions to assess its accuracy and fidelity.
Specifically, sinusoidal power signals were employed as input stimuli for model validation. In EXP\#1 and EXP\#2, a single TCA was actuated following a sinusoidal power input with frequencies of 1/60\,Hz and 1/120\,Hz, respectively. In EXP\#3, two TCAs were actuated, each following a sinusoidal power input with a frequency of 1/120\,Hz and a phase shift of 180$^\circ$ between them. EXP\#4 involved all three TCAs actuated at  1/120\,Hz with a 120$^\circ$ phase shift between each pair of control inputs. Figure~\ref{sinmodelval} illustrates the model validation experimental results for EXP\#1--\#4. We observe mean absolute errors (MAEs) of 0.34$^\circ$ and 0.66$^\circ$ in the bending angle for EXP\#1 and EXP\#2, respectively, accounting for 2.3$\%$ and 4.1$\%$ of the maximum amplitude. In the multi-input cases, the MAE in the bending angle is 0.42$^\circ$ in EXP\#3, approximately 3.16$\%$ of the maximum amplitude, while the MAE in bending direction is 6.4$^\circ$, accounting for about 3.2$\%$ of the maximum amplitude. In EXP\#4, the MAE in the bending angle is 0.61$^\circ$, approximately 6.6$\%$ of the maximum amplitude, and the MAE in bending direction is 11.6$^\circ$, about 3.2$\%$ of the maximum amplitude.

\par

\begin{table}[htbp]
\begin{center}   
\caption{Prediction Error Between Model Predictions and Experimental Results}
\label{table:model validation} 
\begin{tabular}{|c|c|c|c|c|}   
\hline   \multirow{2}{*}{ } & \multicolumn{2}{|c|}{Bending angle $\theta$} & \multicolumn{2}{|c|}{Bending direction $\varphi$} \\
\cline{2-5}
   & RMSE & MAE & RMSE & MAE \\
\hline   EXP\#1 & 0.35$^\circ$ & 0.34$^\circ$& 0.53$^\circ$ &0.48$^\circ$   \\      
\hline   EXP\#2 & 0.82$^\circ$ &  0.66$^\circ$& 0.27$^\circ$& 0.24$^\circ$\\ 
\hline   EXP\#3 & 0.45$^\circ$ &  0.42$^\circ$& 6.46$^\circ$& 6.18$^\circ$  \\  
\hline   EXP\#4 & 0.69$^\circ$ & 0.61$^\circ$& 13.64$^\circ$ &11.43$^\circ$   \\
\hline 
\hline
$\%$ Error & 4.95$\%$ & 4.01$\%$ & 2.17$\%$ & 1.76$\%$\\
\hline
\end{tabular}   
\end{center}   
\end{table}

Table \ref{table:model validation} presents the root mean square error (RMSE) and the MAE of the prediction errors across all six sets of experiments. Additionally, the percentage errors (\% Error) averaged over all experiments for both RMSE and MAE, with respect to the corresponding amplitudes of movement, are calculated. The experimental results indicate that, in terms of bending angle, the RMSE and MAE of prediction errors account for approximately 7.39$\%$ and 6.07$\%$ of their associated amplitudes of movement, respectively. For the bending direction, the RMSE and MAE of prediction errors are approximately 4.31$\%$ and 2.91$\%$ of their corresponding amplitudes of movement, respectively. In addition, large prediction errors in bending direction occur in EXP\#3 and EXP\#4. We speculate that this is due to the positive correlation between RMSE/MAE and the amplitude of movement.\par

Despite the observed discrepancies between model predictions and experimental results, we consider the accuracy of the established dynamic model satisfactory. This assessment takes into account the unmodeled inherent complexities of the nonlinear TCA dynamics and the inevitable friction present within the parallel mechanism. 
These simplifications represent a deliberate engineering trade-off aimed at preserving the real-time feasibility and computational tractability of the NMPC framework. The closed-loop architecture of the controller  inherently compensates for such unmodeled dynamics, thereby ensuring robust trajectory tracking performance in the presence model-reality mismatches.

\subsection{Comparison Against Other Relevant Designs}

To demonstrate the uniqueness of the proposed robotic wrist, we conducted a comparison against relevant robotic systems reported in the literature, focusing on key metrics such as motion range, modeling approach, control strategy, structural dimension, and maximum actuation frequency. 

\begin{figure}[htbp]
        \centering
        \includegraphics[width=0.5\textwidth]{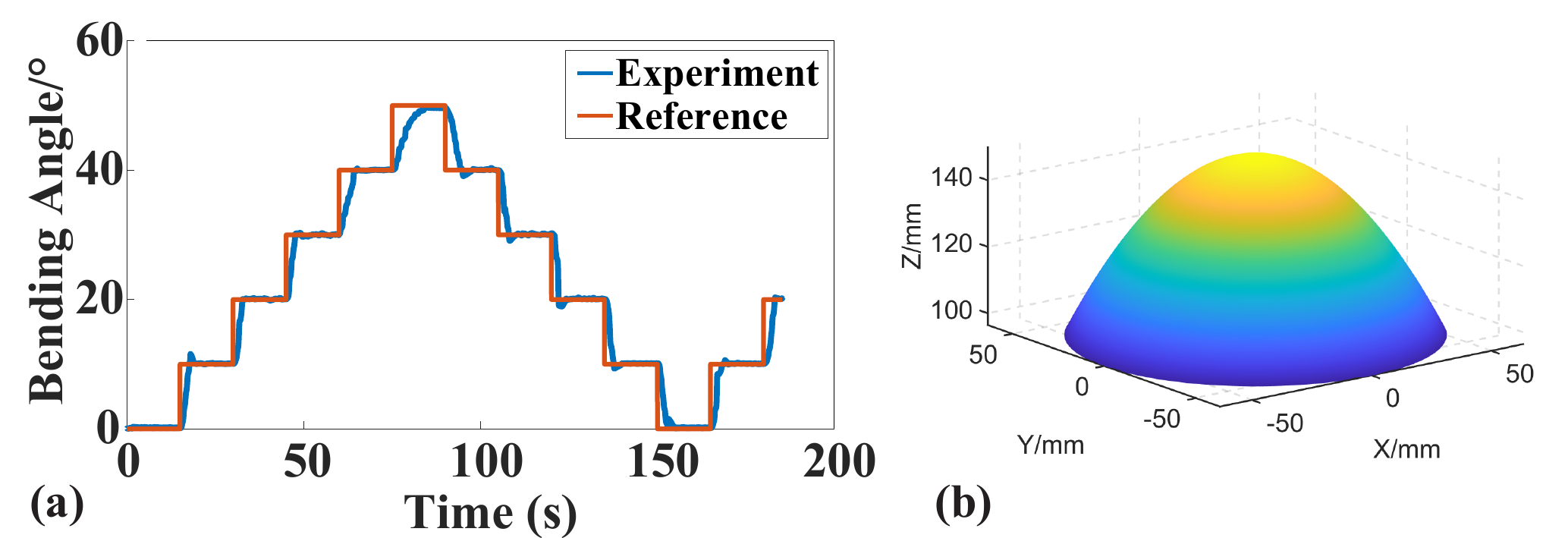}
        \caption{Experimental results demonstrating the motion space of the robotic wrist. (a) Bending angle trajectory $\theta$ during step-wise reference tracking; (b) maximum reachable motion space of the robotic wrist,  illustrating the effective workspace in terms of $\theta$ and $\varphi$.}
        \label{workspace}
\end{figure}

\begin{figure}[htbp]
        \centering
        \includegraphics[width=0.5\textwidth]{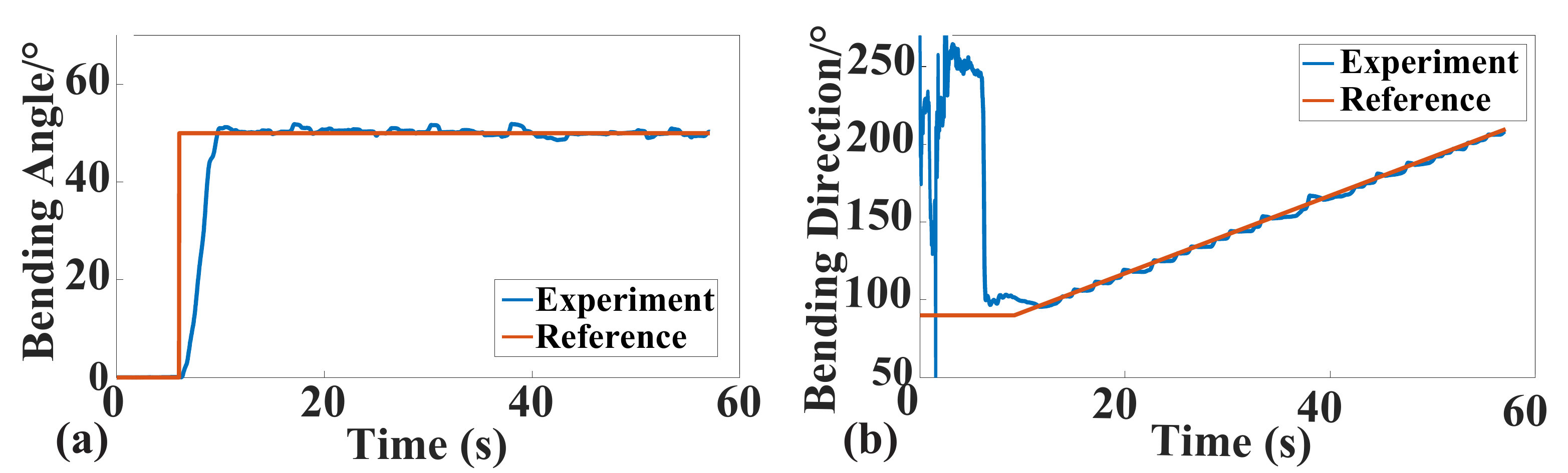}
        \caption{Experimental trajectories of bending angle and bending direction used to evaluate the motion space of the robotic wrist in the bending direction.  (a) Bending angle $\theta$ regulated at a fixed value of 50$^\circ$; (b) bending direction $\varphi$ during controlled tracking from 90$^\circ$ to 210$^\circ$.}
        \label{workspace2}
\end{figure}

To obtain performance benchmarks, a series of experiments were designed and conducted. First, the motion space in the bending angle $\theta$ was evaluated by incrementally increasing $\theta$ in 10$^\circ$ steps every 15 seconds, ranging from 0$^\circ$ to 50$^\circ$, while maintaining a fixed bending direction $\varphi = 90^\circ$. As illustrated in Fig.~\ref{workspace}(a), precise actuation was sustained up to 50$^\circ$, beyond which control performance degraded. To assess the motion range of $\varphi$, we conducted a second experiment. After setting the wrist to a bending angle of $\theta = 50^\circ$, we gradually adjusted the bending direction $\varphi$ from 90$^\circ$ to 210$^\circ$, as illustrated in Fig.~\ref{workspace2}. Considering the wrist's 120$^\circ$ structural symmetry, the effective workspace spans $\theta$ from 0$^\circ$ to 50$^\circ$ and $\varphi$ from 0$^\circ$ to 360$^\circ$, as summarized in Fig.~\ref{workspace}(b). This characterization confirms the wrist's capability for full planar directional control within its bending range, thereby establishing its suitability for versatile spatial manipulation tasks.

\begin{table*}[b]
\centering
\caption{Comparison Against Other Relevant Designs}
\label{robot_compar1}
\begin{tabular}{|*{8}{c|}}
\hline
Robot & Actuator & Motion& Model & Control & Dimension   & Frequency & Max  \\
& Mode & Range&  Approach  & Strategy &(cm)     & (Hz) &Load(kg) \\
\hline
Our work & TCA & 50$^\circ$& Dynamic & NMPC & 17*9   & 0.04 & 0.2 \\
\hline
MISR\cite{sun2023development}& TCA  & 92$^\circ$& Kinematic  & PI & 31.2*11  & 0.002 & N/A  \\
\hline
Arm\cite{yang2020compact}& TCA & 14$^\circ$& Kinematic& PID & N/A & N/A  & N/A \\
\hline
SRM \cite{yang2024variable}& TCA-SMA  &76$^\circ$& Kinematic & PID &13*3.3   & 0.08 &0.15\\
\hline
MS\cite{wu2018biorobotic}& TCA  & 24$^\circ$& N/A & N/A & 4.5*3  & 0.03 & N/A \\
\hline
Neck\cite{copaci2020sma} & SMA &40$^\circ$& Kinematic  & BPID & 10*9  & 0.025 & 1\\
\hline
MISR\cite{shao2020design} & SMA & 80$^\circ$& Kinematic & PI  with feedforward &305 *76  &  0.01 &N/A \\

\hline
\end{tabular}
\end{table*}

\begin{figure}[htbp]
        \centering
        \includegraphics[width=0.5\textwidth]{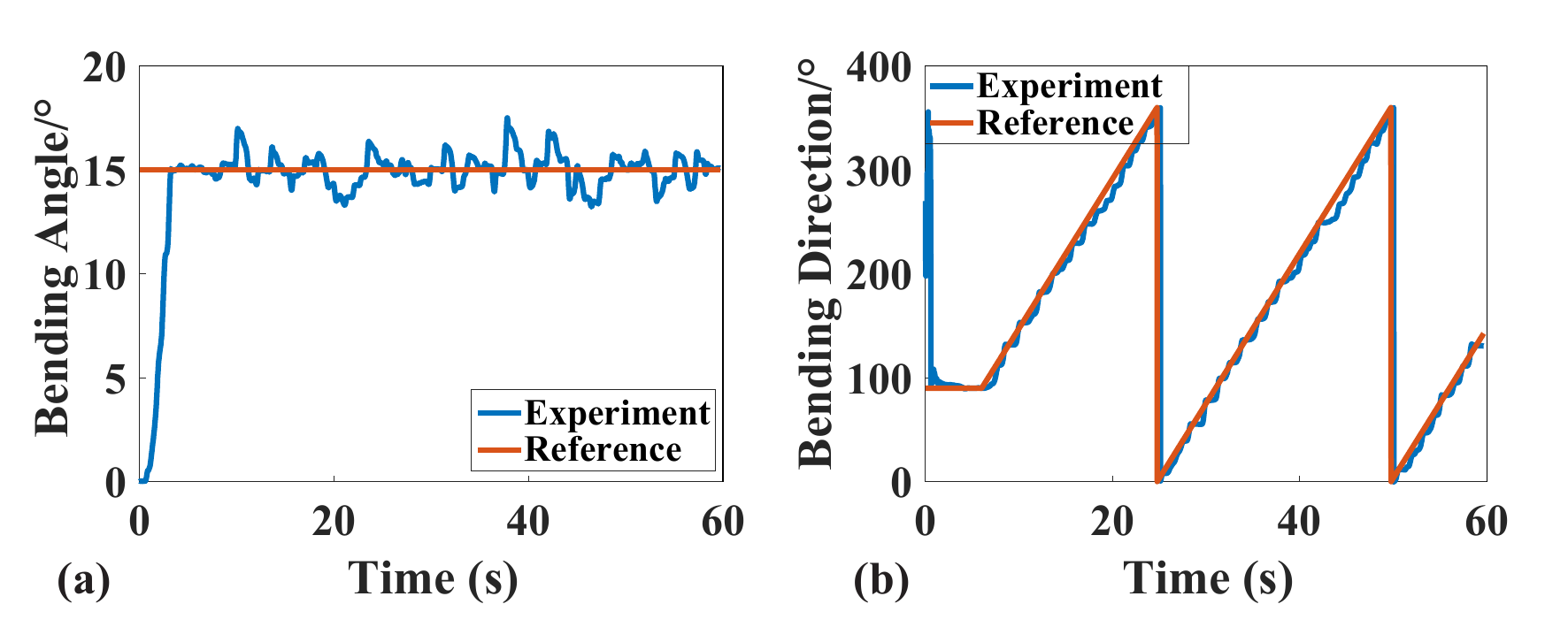}
        \caption{Experimental results of tracking a 0.04\,Hz circular trajectory. (a) Bending angle $\theta$ trajectory; (b) bending direction $\varphi$ trajectory. }
        \label{bandwidth}
\end{figure}

The actuation frequency in trajectory tracking tasks was evaluated by setting a repetitive reference trajectory and observing the robotic wrist's corresponding motion response. Following the methodology adopted in prior work on TCA-driven minimally invasive surgical robots (MISR) \cite{sun2023development}, we selected a fixed circular trajectory, defined by a constant bending angle $\theta = 15^\circ$ and a time-varying bending direction $\varphi = 2\pi f t$, where $f$ and $t$ represent the input frequency and time, respectively. Experimental results in Fig.~\ref{bandwidth} illustrate accurate closed-loop tracking at 0.04 Hz, identified as the maximum achievable operating frequency under the current control scheme. A comparative evaluation of the robotic wrist against existing TCA-based designs is summarized in Table~\ref{robot_compar1}. In terms of motion range, the proposed wrist ranks in the upper-middle tier, second only to the MISR by Sun \textit{et al}. \cite{sun2023development} and the SRM by Yang \textit{et al.} \cite{yang2024variable}. With a maximum actuation frequency of 0.04,Hz, it also ranks second, following the SRM. Overall, the proposed robotic wrist exhibits strong performance across all comparison metrics. Notably, most prior works did not incorporate comprehensive dynamic modeling or advanced control methodologies, relying primarily on simplified kinematic models and PID controllers.

\subsection{Trajectory Tracking of Robotic Wrist}
\label{track_sec}

The NMPC framework was implemented in MATLAB, utilizing the Model Predictive Control Toolbox and the \textit{fmincon} solver with the sequential quadratic programming (SQP) algorithm. For performance benchmarking, we designed a proportional-integral-derivative (PID) controller with carefully tuned parameters $K_p = 48$,  $K_i=0.05$ and $K_d = 7.2$, ensuring a fair and robust baseline. The input to the PID was defined as the Euclidean norm of the difference between two TCA length vectors---one calculated from the desired bending angle and direction, and the other obtained from IMU measurements via the geometric relations in Eqs.~\eqref{4} - \eqref{6}. The controller output specifies the power inputs required to actuate the TCAs. Both NMPC and PID controllers operated at a fixed control frequency of 10\,Hz. To validate the effectiveness of the NMPC, a series of experiments involving complex trajectories under varied environmental conditions were conducted.

\begin{figure*}[t]
        \centering
        \includegraphics[width=0.85\textwidth]{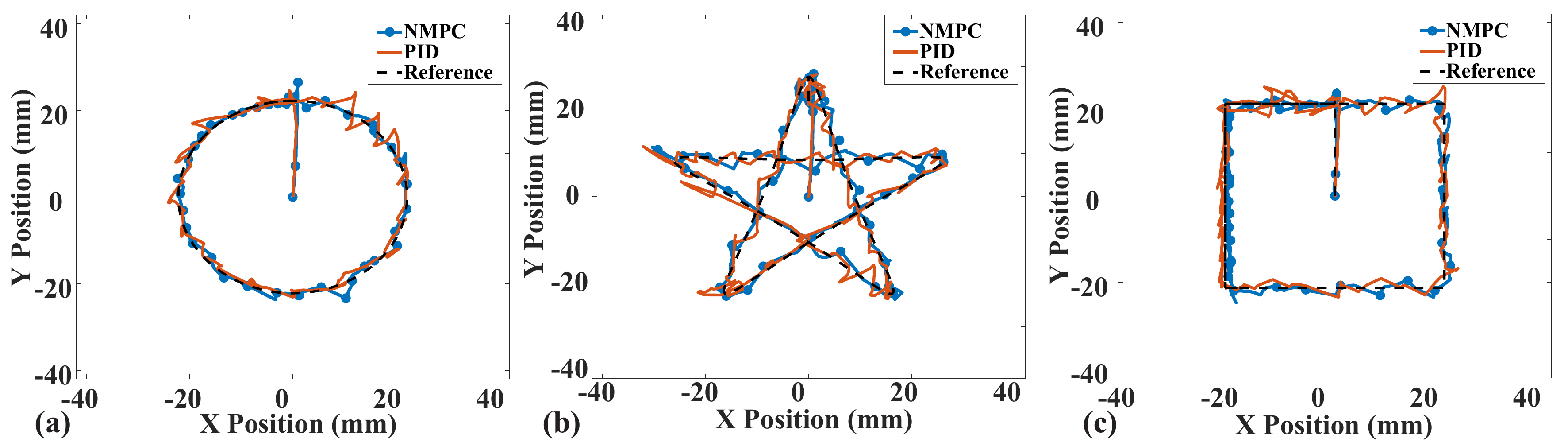}
        \caption{Top view comparison of NMPC and PID control results from the second set of experiments. (a) Circular trajectory tracking; (b) star-shaped trajectory tracking, and (c) square-shaped trajectory tracking.}
        \label{track_second_top}
\end{figure*}

\begin{figure*}[h]
        \centering
        \includegraphics[width=1\textwidth]{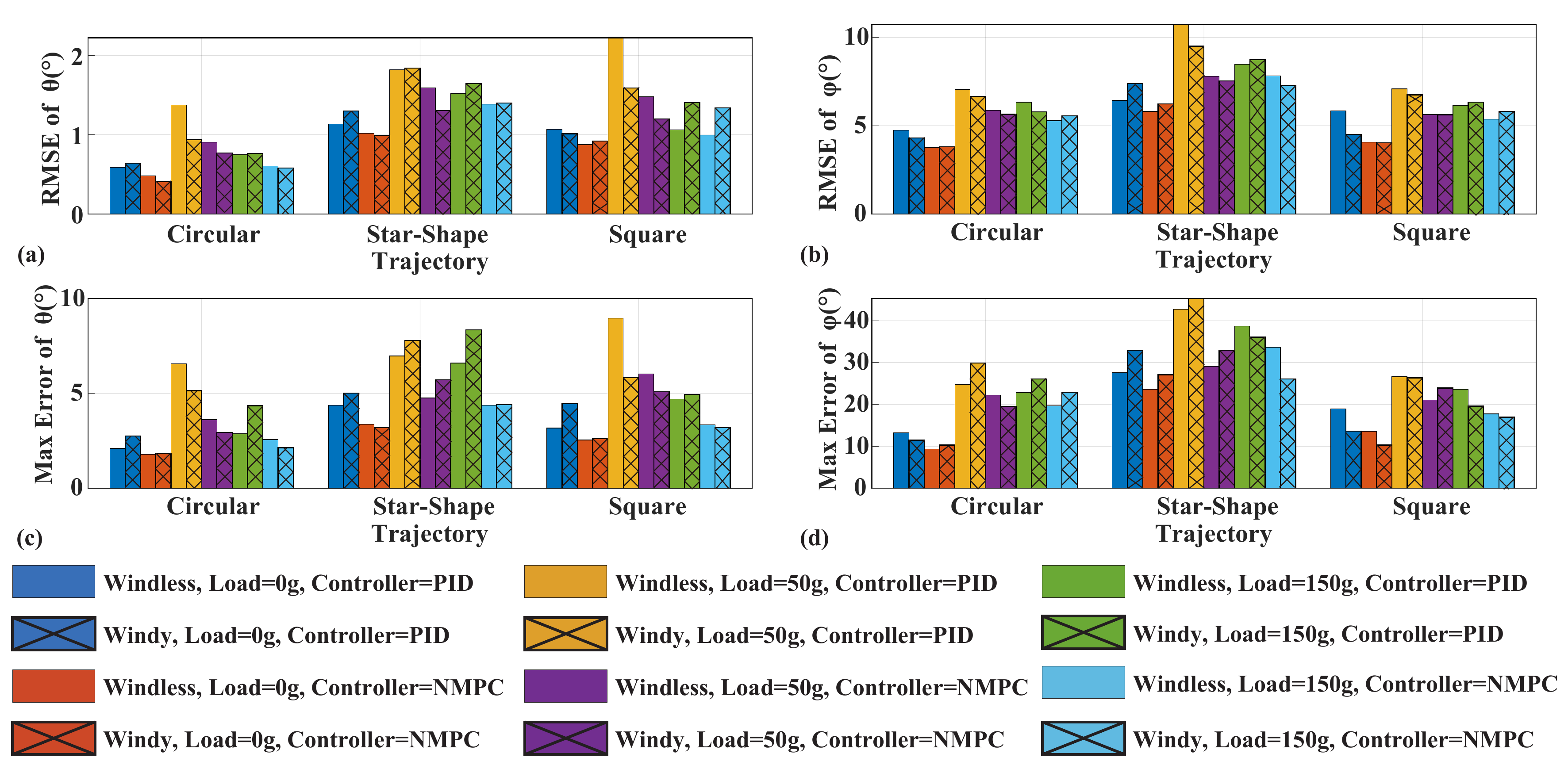}
        \caption{Experimental results of the robotic wrist controlled under NMPC and PID control across varying thermal conductivity, reference trajectories, and load configurations. (a) RMSE of bending angle $\theta$; (b) RMSE of bending direction $\varphi$; (c) ME of bending angle $\theta$; and (d) ME of bending direction $\varphi$.}
        \label{windbarchart}
\end{figure*}

To evaluate the control performance of NMPC under varying load conditions, three reference trajectories---a circle, a square, and a star---were selected with periods of 60\,s, 60\,s, and 90\,s, respectively. Loads of 0\,g, 50\,g, and 150\,g were applied to the robot's end-effector. For the 150\,g load condition, two TCAs were connected in parallel at the TCA placement position on the robotic wrist to ensure adequate load management. The selection of these trajectories was intentional to assess different aspects of control performance: the circular trajectory, characterized by smooth and continuously differentiable motion, tested the controller's steady-state tracking capability; the square trajectory, with its sharp 90-degree corners, challenged the controller's responsiveness to abrupt velocity changes; and the star-shaped trajectory, incorporating both sharp angles and smooth segments, served as comprehensive benchmark for overall  controller's robustness across complex geometries. 

Fig.~\ref{track_second_top} illustrates the top-view tracking results of the robotic wrist executing three reference trajectories under no-load conditions using both NMPC and PID controllers. For trajectories with shorter periods, the PID controller exhibited notable tracking fluctuations, particularly at sharp corners of the square and star trajectories. In contrast, the NMPC controller maintained higher fidelity across all cases. Table~\ref{track error} quantifies the tracking errors under this no-load condition. We observe that the NMPC reduced RMSE and Maximum Error (ME) of the bending angle $\theta$ by 17.4$\%$ and 20.1$\%$, respectively, compared to the PID controller. In terms of the bending direction $\varphi$,  NMPC achieved error reductions of 20.3$\%$ in RMSE and 24.1$\%$ in ME. The performance advantage of NMPC becomes even more evident under payload conditions, as illustrated in Fig.~\ref{windbarchart}, which compares the RMSE and ME for both controllers across all test cases. While the tracking errors for both controllers increase with added load, NMPC consistently outperformed PID. Averaged across all non-zero load conditions, the NMPC controller reduced the RMSE and ME of the bending angle $\theta$ by 17.9\% and 26.7\%, respectively. For the bending direction $\varphi$, the average reductions in RMSE and ME were 18.1\% and 20.8\%, respectively. These results confirm that the proposed NMPC framework offers significantly improved tracking accuracy and robustness to payload variations compared to conventional PID control.

\begin{table}[htbp]
\centering
\caption{The experimental results of RMSE and ME for NMPC and PID in trajectory tracking tasks under no-load conditions.}
\label{track error} 
\begin{tabular}{c c cc cc}
\toprule
Controller &
Reference &
\multicolumn{2}{c}{Bending angle $\boldsymbol{\theta}$($^\circ$)} &
\multicolumn{2}{c}{Bending direction $\boldsymbol{\varphi}$($^\circ$)} \\
\cmidrule(lr){3-4} \cmidrule(lr){5-6}
Type &
Trajectory & RMSE & ME & RMSE & ME \\
\midrule
\multirow{3}{*}{NMPC}
& $\bigcirc$ &  0.48 & 1.79 & 3.77 &  9.34 \\
& $\mytikzstar$    &  1.02 & 3.37 & 5.81 &  23.58  \\
& $\square$    &  0.87 & 2.54 & 4.06 &  13.55 \\
\midrule
\multirow{3}{*}{PID}  
& $\bigcirc$ & 0.59 & 2.10 & 4.75 & 13.21 \\
& $\mytikzstar$    & 1.20 & 4.37 & 6.44 & 27.63  \\
& $ \square$    & 1.07 & 3.3 & 5.84 & 18.94  \\
\midrule
\multicolumn{2}{c}{Average Improvement} & 17.4\% & 20.1\% & 20.3\% & 24.1\%  \\
\bottomrule
\end{tabular}
\end{table}

In the second set of experiments, we evaluated the robustness of NMPC to environmental variations by introducing airflow to increase the thermal conductivity $\lambda$ of the TCA. All other experimental settings, including reference trajectories and applied loads, remained consistent with the first set of experiments. Fig.~\ref{windbarchart} presents the RMSE and ME for trajectory tracking under windy conditions. The airflow improved the TCA's thermal dissipation,, resulting in a faster dynamic response and a reduction in tracking errors. Across all evaluated conditions, NMPC consistently outperformed the PID controller. Specifically, for the bending angle $\theta$, NMPC reduced RMSE by 20.4\% and ME by 36.4\% on average. For the bending direction $\varphi$, RMSE and ME reductions averaged 13.2\% and 19.6\%, respectively. These results confirm that NMPC maintains high tracking accuracy and robustness under varying thermal conductivity $\lambda$, demonstrating strong adaptability to environmental changes.

Under varying thermal conductivity and load conditions across all reference trajectories, NMPC consistently exhibited superior motion control precision and enhanced robustness compared to the PID controller. Quantitatively, NMPC achieved  an overall improvement of approximately 20$\%$ in trajectory tracking accuracy for both the bending angle $\theta$ and the bending direction $\varphi$, highlighting controllers grounded in dynamic modeling significantly outperform traditional PID approaches in the motion control of TCA-driven robotic systems.

\section{Robotic System Demonstration Using Modular TCA-Driven Wrists}

To further investigate the practicality and scalability of our design as a modular building block, we developed and experimentally evaluated a Multi-Segment Soft Robot Arm (MSRA) composed of multiple miniaturized robotic wrist.

First, we developed and tested a miniaturized version of the wrist module,  featuring a compact form and reduced mass. Following a methodology similar to that outlined in earlier sections, we evaluated its range of motion and motion bandwidth. Key performance parameters are summarized in Table~\ref{mini_wrist}. 

To assess the controller's portability, we identified the dynamic parameters of the miniaturized wrist and implemented the NMPC controller for this scaled-down system.  Its trajectory tracking performance was compared against against a PID controller using the same circular, square, and star-shaped reference trajectories described in Section~\ref{track_sec}, with periods of 60~s, 60~s, and 120~s, respectively. The results presented in Table~\ref{mini_wrist} indicate that NMPC reduced the RMSE in tracking the bending angle and orientation by 21.1\% and 19.4\%, respectively, relative to the PID controller. This performance enhancement is highly consistent with the results from the full-scale wrist, demonstrating the portability and scalability of our proposed NMPC strategy across different robot sizes.

\begin{table}[htbp]
    \captionsetup{justification=centering, singlelinecheck=false}
    \centering
    \caption{Performance Evaluation of the Miniaturized Robotic Wrist and its Controller}
    \label{mini_wrist}
    \begin{tabular}{
        @{}
        l
        l
        S[table-format=3.2]
        @{}
    }
        \toprule
        \textbf{Performance Metric} & & {\textbf{Value}} \\ 
        \midrule
        \multicolumn{3}{l}{\textbf{Kinematic \& Physical Properties}} \\
        \addlinespace
        
        \quad Motion Range  & & \SI{40}{\degree} \\
        \quad Motion Bandwidth  &    & \SI{0.03}{\hertz} \\
        \quad Dimensions (L $\times$ W) & &{70\,\unit{\milli\meter} $\times$ 45\,\unit{\milli\meter}} \\
        \quad Mass & & \SI{58.5}{\gram} \\
        \addlinespace
        
        \multicolumn{3}{l}{\textbf{Controller Performance }} \\
        \addlinespace
        
        \quad Bending Angle RMSE Improvement &    & \SI{21.1}{\%} \\
        \quad Bending Direction RMSE Improvement& & \SI{19.4}{\%} \\
        \bottomrule
    \end{tabular}
\end{table}

\begin{figure}[htbp]
        \centering
        \includegraphics[width=0.45\textwidth]{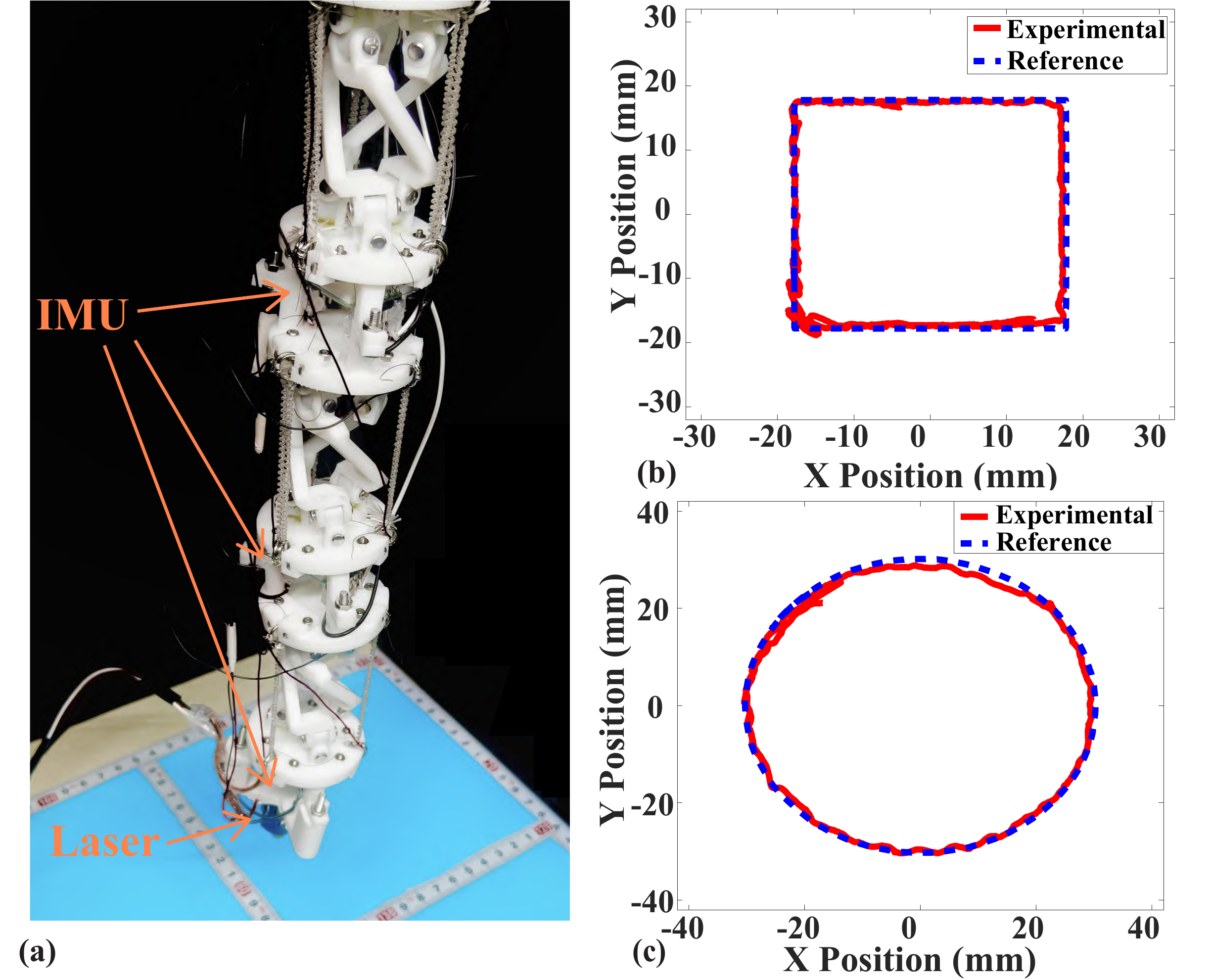}
        \caption{Demonstration of the MSRA and its trajectory tracking performance. (a) The MSRA prototype, constructed by serially connecting three miniaturized wrist modules. (b) Top view of the square trajectory tracking. (c) Top view of the circular trajectory tracking.}
        \label{demo2}
\end{figure}

To further demonstrate modular scalability, we constructed a full MSRA by serially connecting three miniaturized wrist modules. Compared to conventional electric motor or pneumatic actuation schemes, the use of TCAs offer a more compact and lightweight architecture. The entire MSRA weighs only 220\,g and provides six DoF. To accommodate segment-specific torque demands, a hierarchical TCA configuration was implemented: the top, middle, and bottom modules were equipped with three, two, and one TCA per side, respectively, as depicted in Fig.~\ref{demo2}(a). The control system for the MSRA remained largely identical to that of a single wrist module, with the primary modification being the inclusion of additional sensors and electronic components. For preliminary validation, PID control was used to command square and circular trajectories with 200 s and 180 s periods. As shown in Fig.~\ref{demo2}(b) and (c), the MSRA achieved excellent trajectory tracking with acceptable errors. These results confirm the feasibility of using our wrist module as a fundamental building block for more complex robotic systems.

For trajectories with shorter periods and larger ranges, the MSRA exhibited notable tracking errors. We conjecture that these are attributed to strong inter-segment coupling, error accumulation along the arm due to increased DoF, and the limitations of simple PID control. Enhancing tracking accuracy requires a more accurate system model and advanced control strategies, as supported by recent studies \cite{weissman2025efficient, chen2025versatile}. While modeling the full MSRA is highly challenging, the single-module model presented here offers a solid theoretical basis and a starting point for future developments.

\section{Conclusion}

This paper proposed a 2-DOF robotic wrist module actuated by twisted and coiled actuators (TCAs). A Lagrangian dynamic model of the robotic wrist was derived, addressing a gap in the existing literature on TCA-driven parallel wrists. Based on this model, a nonlinear model predictive controller (NMPC) was developed for trajectory tracking. Extensive experiments were conducted, the results of which validated the established dynamics model. Furthermore, systematic comparisons between NMPC and PID control under various trajectories and conditions consistently demonstrated the tracking accuracy and robustness of the NMPC design.To demonstrate its practical integration, the wrist module was miniaturized and employed as a building block within a modular soft robotic arm (MSRA), successfully completing a trajectory tracking task.

In future work, we plan to explore the deployment of the proposed TCA-driven robotic wrists in more complex systems, such as MSRAs, leveraging advanced control algorithms to enable practical applications. To address the limited actuation speed of TCAs caused by thermal cycling, we will integrate actively controlled fans to enhance response time. Additionally, alternative model-based control strategies will be investigated to improve precision and overall control performance.

\small


\bibliographystyle{IEEEtran}
\bibliography{ref.bib}

\normalsize 
\begin{IEEEbiography}[{\includegraphics[width=1in,height=1.25in,clip,keepaspectratio]{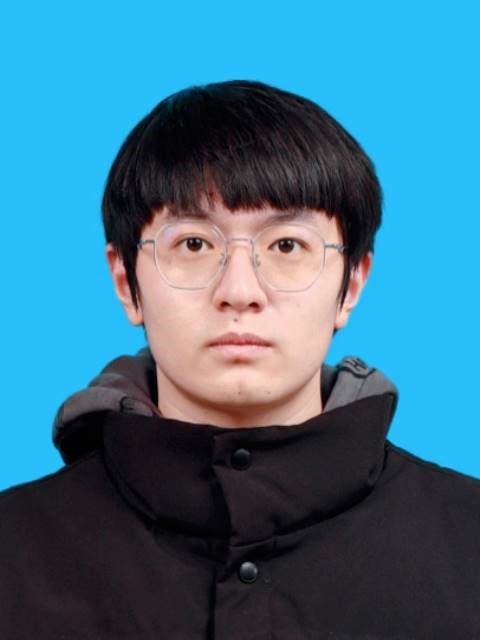}}]{Yunsong Zhang}
 received his Bachelor's degree in Robotics Engineering from College of Engineering at Peking University, Beijing, China in 2024. He is currently a Ph.D. student in general mechanics and foundation of mechanics with the College of Engineering at Peking University, Beijing, China in 2024. His current research interests include bio-inspired robotics and artificial muscles.
\end{IEEEbiography}
\begin{IEEEbiography}
[{\includegraphics[width=1in,height=1.25in,clip,keepaspectratio]{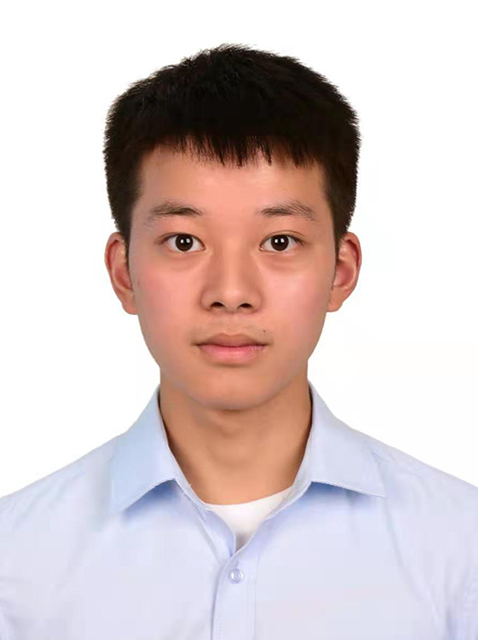}}]{Xinyu Zhou}
received his Bachelor's degree in Robotics Engineering from the College of Engineering at Peking University, Beijing, China in 2023. He is currently a Ph.D. student in the Department of Electrical and Computer Engineering at Michigan State University, East Lansing, MI, USA. His current research interests include soft robots and artificial muscles.
\end{IEEEbiography}
\begin{IEEEbiography}[{\includegraphics[width=1in,height=1.25in,clip,keepaspectratio]{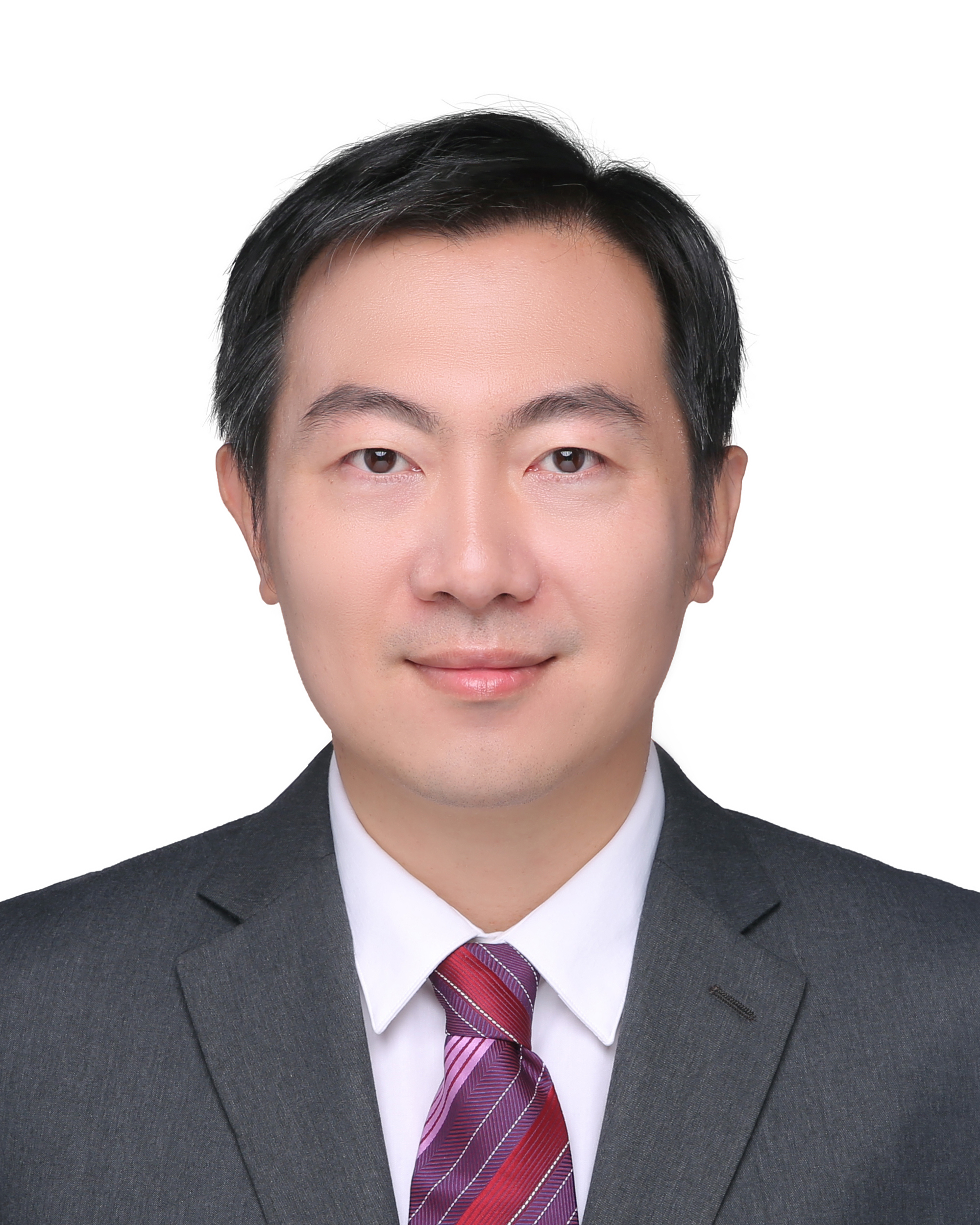}}]{Feitian Zhang}(S'12–M'14) received the Bachelor's and Master's degrees in automatic control from Harbin Institute of Technology, Harbin, China, in 2007 and 2009, respectively, and the Ph.D. degree in electrical and computer engineering from Michigan State University, East Lansing, MI, USA, in 2014. He was a Postdoctoral Research Associate with the Department of Aerospace Engineering and Institute for Systems Research at University of Maryland, College Park, MD, USA from 2014 to 2016, and an Assistant Professor of Electrical and Computer Engineering with George Mason University, Fairfax, VA, USA from 2016 to 2021. He is currently an Associate Professor of Robotics Engineering with Peking University, Beijing, China. His research interests include mechatronics systems, robotics and controls, aerial vehicles and underwater vehicles.
\end{IEEEbiography}
\vfill
\newpage

\end{document}